\PassOptionsToPackage{dvipsnames}{xcolor}
\documentclass{article}
\usepackage{spconf,amsmath,graphicx}
\usepackage{times}
\usepackage{latexsym}
\usepackage[utf8]{inputenc}
\usepackage{microtype}
\usepackage{soul}
\usepackage{url}
\usepackage{color}
\usepackage[dvipsnames]{xcolor}
\usepackage{multirow}
\usepackage{graphicx}
\usepackage{amsmath}
\usepackage{amsthm}
\usepackage{booktabs}
\usepackage{algorithm}
\usepackage{algorithmic}
\usepackage{CJKutf8}
\usepackage{amsfonts}
\usepackage{amssymb}
\usepackage{bbm}

\title{CENTRE: A PARAGRAPH-LEVEL CHINESE DATASET FOR RELATION EXTRACTION AMONG ENTERPRISES}
\name{Peipei Liu$^{1,2}$,Hong Li$^{1,2}$\thanks{Corresponding Author: Hong Li lihong@iie.ac.cn},Zhiyu Wang$^{3}$,Yimo Ren$^{1,2}$,Jie Liu$^{1,2}$,Fei Lyu$^{2}$,Hongsong Zhu$^{1,2}$,Limin Sun$^{1,2}$ }
\address{$^{1}$ School of Cyber Security, University of Chinese Academy of Sciences \\ $^{2}$ Institute of Information Engineering, Chinese Academy of Sciences \\ $^{3}$School of Software, Henan University}
\begin{document}
%
\maketitle
\begin{abstract}
  Enterprise relation extraction aims to detect pairs of enterprise entities and identify the business relations between them from unstructured or semi-structured text data, and it is crucial for several real-world applications such as risk analysis, rating research and supply chain security. However, previous work mainly focuses on getting attribute information about enterprises like personnel and corporate business, and pays little attention to enterprise relation extraction. To encourage further progress in the research, we introduce the CEntRE, a new dataset constructed from publicly available business news data with careful human annotation and intelligent data processing. Extensive experiments on CEntRE with six excellent models demonstrate the challenges of our proposed dataset.
\end{abstract}
\section{Introduction}
\label{sec:intro}

Enterprise relation extraction aims to automatically detect pairs of enterprise entities and identify the business relations between them from unstructured or semi-structured text data without human intervention \cite{12}, and it is a sub-task in the natural language processing (NLP). In the real world, enterprise relation extraction plays an important role in several fields such as economics and finance, information security and supply chain security, etc. On one hand, invest institutions can get the relations of equity distribution and debtor-creditor among different enterprises based on the relation extraction, and these relations are then used for risk analysis and rating research. On the other hand, as an important part of enterprise operation, the supply chain relations can help enterprise understand and analyze industries, make management decisions and select business partners, as well as enhancing the competitiveness and improving profit margin. More than that, the analysis of enterprise relations referring to social engineering is beneficial to effective surveillance and precautions for network penetration since the low security protection of subsidiary enterprises may threaten the superior enterprises and the partnership is easy to be exploited by phishing attacks, etc. Therefore, relation extraction among enterprises is crucial to the security of development and reduction of economic losses. 

Previous work mainly focuses on getting attribute information about enterprises like personnel, corporate business, product or certain relations (e.g., chip supply chain) instead of general relations among enterprises from unsupervised data by using rules, dictionary, template and other hand-craft features\cite{YangCL,MengL,DaiJB,SunC}. Despite desirable results could be achieved, these content and methods are trapped by some inevitable restrictions in practice: they are applied to domain-specific, and are quite difficult to be used for conducting business analysis comprehensively enough; with the explosion growth and the increasing diversity of new text data, fixed patterns and manual features become more expensive and unapplicable.

Recent developments in deep learning have promoted the research of neural relation extraction (NRE), which attempts to use neural networks to automatically learn high-level semantic features and then extracts the entity relations\cite{12,13,14,39,41,ChenDQ,Dai,TwoThanOne}. The models trained with neural networks are conveniently fine-tuned for various types of content and knowledge, which is conducive to the migration and application of extraction methods. Although NRE brings ideas to the further research about enterprise relations, deep learning is data-consuming and there is still a lack of unified gold-standard supervised dataset for enterprise relation extraction which hinders the future research in this area.

According to the above analysis, we take an initial step towards studying relation extraction (RE) among enterprises based on deep learning in this paper.

\begin{figure*}[h]
  \centering
  \includegraphics[height=1.4in,width=5.8in]{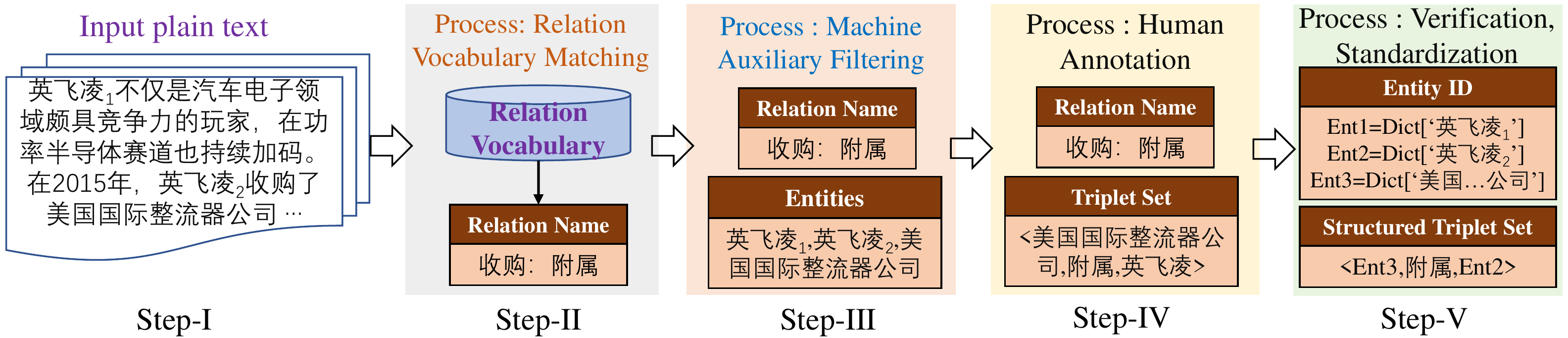}
  \caption{The data selection and annotation process}  
  \label{data_process}
  \vspace*{-0.4cm}
\end{figure*}

Firstly, a Chinese relation extraction dataset called CEntRE is constructed. Specifically, the dataset contains 11018 relational triples and 10894 enterprise entities on 3018 paragraphs, and 19 relations that exist between pairs of enterprise entities are covered. The CEntRE has following four characteristics: (I) Overlapping entity: The relational facts in paragraphs of CEntRE are often complicated, and different relational triplets may have overlaps in the same entities. (II) Flexible name structure: Unlike the registration information in the State Administration for Market Regulation, the enterprise names in CEntRE are arbitrary and easy to be confused with other common words, such as \begin{CJK*}{UTF8}{gbsn}“我爱网络公司” (I Love Network Company) and “交个朋友” (Make Friends)\end{CJK*}. (III) Cross-sentence and Synonyms reasoning: As a substantial portion of the relational facts, some relations in CEntRE can only be extracted from multiple sentences. It means that CEntRE requires reading multiple sentences in a paragraph to recognize entity pairs and inferring their relations by synthesizing all information of the paragraph. In addition, Chinese synonyms should not be ignored, such as \begin{CJK*}{UTF8}{gbsn}
“销往” “供给” and “供应”\end{CJK*}. (IV) Wide coverage: The CEntRE originates from more than 20000 news media reports among 7 websites of technology and finance in past 6 months, and those abstract enterprise relations can cover most of the application scenarios in real world.

Then, we analyze the characteristics of CEntRE, and have experiments on the CentRE with several excellent neural networks models for enterprise entity recognition and relation extraction to demonstrate its challenges. 

In summary, this paper makes following contributions: 
\begin{itemize}
\item A labeled Chinese enterprise relation extraction dataset CEntRE with 3018 paragraphs and 19 relations is released for analyzing the interactions among the enterprises. The dataset is at paragraph-level, and it has some particular challenges.
\item We conduct extensive evaluation on six state-of-the-art NRE methods in various setting. Experiment results show the challenges of our dataset.
\item Furthermore, detailed analysis on the results also reveals multiple promising directions worth pursuing.
\end{itemize}

\section{Related Works}
\label{sec:related_work}
The values of enterprise relation management are receiving attention from researchers and practitioners in several sectors with the growing market world economy. 

\cite{YangCL} concerns on the supply relation between enterprises, especially the role of corporate entity. A library of relation word is built to judge the theme of the text, and the nearest syntactic dependent verbs are used to judge the semantic relation between enterprise entities. By using the maximum entropy model to analyze the unlabeled data and determine the optimal feature template, \cite{SunC} constructs the enterprise knowledge graph for financial application. \cite{MengL} presents a pipeline model for company name identification and relation extraction based on fusing rules, dictionary matching and machine learning algorithm. However, these methods cost a lot of manual efforts and are difficult to be generalized. 

\cite{DaiJB} introduces the triggering mechanism, and then the model with triggering words constraint is used to extract 5 types of relation from the standard annual report of Chinese Listed Firms. Although 
the method can obtain good results, it is limited to standard data. \cite{8} designs automatic models to get relations and entities about enterprises, whereas they mainly focused on the information of enterprise-self such as location, personnel and industry. FinRE \cite{11} is the closest work to ours, but it is simpler with only two enterprise entities and one relation between them in a single sentence. 

\begin{CJK*}{UTF8}{gbsn}
  \begin{table*}[htbp]
  \scriptsize
  \centering
  \begin{tabular}{|c|c|l|c|c|}
  \cline{1-5}
  Chinese & English                                                                                  & \multicolumn{1}{c|}{Description}                                                                                                                                                                             & Count & Symmetric \\ \cline{1-5}
  合作      & Cooperation                                                                              & \begin{tabular}[c]{@{}l@{}}Subject A cooperates with object B. "Cooperation" relationship \\includes signing, win the bidding, customer,etc.\end{tabular}                           & 1184 & True \\ \cline{1-5}
  供应      & Supply                                                                                   & \begin{tabular}[c]{@{}l@{}}Subject A supplys products, businesss, technology,etc. to object B. "Supply" \\relationship includes sold to, procurement, supply, buy, etc.\end{tabular} &629  & False \\ \cline{1-5}
  参股      & \begin{tabular}[c]{@{}c@{}}shareholding/joint-stock/\\ equity participation\end{tabular} & \begin{tabular}[c]{@{}l@{}}Subject A holds shares in subject B,  or purchases shares in B from a third party.\end{tabular}                                                         & 1162 & False \\ \cline{1-5}
  转让      & transfer                                                                                 & Subject A transfers shares, patents, technologies, etc. to subject B.                                                                                                                & 335 & False \\ \cline{1-5}
  控股      & hold controlling stakes                                                                                  & \begin{tabular}[c]{@{}l@{}}Subject A holding object B.  Holding and shareholding are essentially the same, \\except for different explicit expressions and shares.\end{tabular}      & 470 & False \\ \cline{1-5}
  附属      & be the subsidiary of/belong to                                                                               & Subject A is owned by object B.                                                                                                                                                      & 2095 & False \\ \cline{1-5}
  合资      & joint venture                                                                            & Subject A and object B co-create C.                                                                                                                                      &  241& True \\ \cline{1-5}
  投资      & investment                                                                               & Subject A invests in object B, but is not sure whether to take a stake                                                                                                               & 819 & False \\ \cline{1-5}
  授权      & licensing                                                                                & Subject A licenses patens, technologies, etc. to object B.                                                                                                                           & 39 & False \\ \cline{1-5}
  代管      & manage on behalf of                                                                      & Subject A entrusts object B to manage certain content on its behalf.                                                                                                                 & 26 & False \\ \cline{1-5}
  合并      & merger                                                                                   & Subject A and object B are reorganized and merged into C.                                                                                                                & 86 & True \\ \cline{1-5}
  剥离      & spin-off                                                                                 & Subject A spins off object B.                                                                                                                                                        & 316 & False \\ \cline{1-5}
  竞争      & competition                                                                              & Subject A competes with object B in certain areas.                                                                                                                                   & 1142 & True \\ \cline{1-5}
  代工      & OEM                                                                                      & Subject A performs OEM production for object B.                                                                                                                                      & 63 & False \\ \cline{1-5}
  委托      & commission                                                                               & Subject A commissions object B to do some work.                                                                                                                                      & 38 & False \\ \cline{1-5}
  更名      & change name/ rename                                                                      & Subject A changed its name to object B.                                                                                                                                              & 390 & False \\ \cline{1-5}
  共指     & abbreviation                                                                             & Object B, short for Subject A. Or Chinese with English.                                                                                                                                                     & 1446 & True \\ \cline{1-5}
  纠纷      & lawsuit                                                                                  & Subject A sues object B for a certain content.  Or commercial contradiction                                   .                                                                                                 & 121 & True \\ \cline{1-5}
  关联      & correlation                                                                              &                                                                                          Subject A is related to Object B, but it is not really a specific relation.                                                                                                                   & 416 & True \\ \cline{1-5}
  \end{tabular}
  \caption{The descriptions of enterprise relations}
  \label{tab:one}
  \vspace*{-0.4cm}
  \end{table*}

  \end{CJK*}

\section{Dataset Construction}
\label{sec:construction}

\subsection{Collection and Annotation}

Our ultimate goal is to construct a dataset for paragraph-level RE from open text, which requires necessary information including enterprise named entities and relations of all entity pairs. Due to the complexity and diversity of enterprise relations, building dataset is time-consuming and difficult. Fig. \ref{data_process} shows the annotation process and we will introduce the processing details in the following:

\textbf{Step1-Relation Dictionary Establishment:} The first asset of high-quality dataset should be a high-coverage of relations space. We abstract 19 types of enterprise relations by summing up the business knowledge\footnote{This includes business materials (annual reports of listed companies from Shanghai Stock Exchange) and financial news (such as “yicai” and “hexun”)}, referencing existing literatures and discussing with domain experts. Table~\ref{tab:one} gives a detailed description of relations. Although individual difference exists in each form of textual expression, it is found that such relations have internal connections. For example, \begin{CJK*}{UTF8}{gbsn}“参股”(equity participation) covers“控股”(hold controlling stakes)\end{CJK*}. 

\textbf{Step2-Dataset Collection and Machine Filtering:} To increase the diversity of data, we crawl more than 20000 news reports from 7 different websites in 6 months (2020.9-2021.3), which cover finance and technology, and these contents are written by different authors with various styles. Based on the predefined relation dictionary, we first expand the relation vocabulary by synonyms(Table~\ref{tab:one} shows some extended samples). Then, these relation words are aligned to paragraph items to coarsely filter irrelevant paragraphs. Since there should be at least two enterprise entities (subject and object) in each paragraph text, we further discard paragraphs containing fewer than 2 entities. A named entity recognition model trained on MSRA dataset is used for detecting enterprise entities and selecting data . MSRA dataset consists of 3 types of entity tags—“LOC” “PER” “ORG”, but we only need the tag “ORG”.

\textbf{Step3-Human Annotation:} We commit multiple master students to annotate the remaining result from second step manually, and each paragraph is annotated by two individuals separately. However, there may be some difference between the two annotations, a third annotator would like to be asked to merge these differences and make the final decision\footnote{In fact, only about 16\% inconsistent paragraph annotations from the two separate annotators need a third intervention in this stage, which indicates that incorrect labels are limited and the annotation is reliable.}. In the worst case, we have to remove the data which all the annotators cannot reach a consensus on. To provide high-quality annotations, all annotators are well trained that a principled training procedure is adopted and every annotator is required to pass test tasks before annotating the dataset. Apart from aforementioned matters, we transform some relations to make them meet the fact while annotating due to the time-series of business activities. For example, we get\begin{CJK*}{UTF8}{gbsn} “C附属(belong to)B” instead of “A转让(transfer the possession to)C” if “A向B转让了C的100{\%}股权”(A has transferred 100\% equity of C to B)\end{CJK*} is presented because that C no longer belongs to A.

\textbf{Step4-Verification and Standardization:} To evaluate the quality of CEntRE, we randomly selected 180 pieces of data and divided them to four extra reviewers who have known our relation requirements (60 pieces each with an overlap of 20). Each reviewer has to evaluate the correctness of the annotation by classifying each piece of data into \{\textit{Correct}, \textit{Incorrect} or \textit{Uncertain}\}. The inter-reviewer agreement score of \textit{Correct} for the overlap data is 96.25\%, which shows that our annotation is successful. At last, we convert the data from string to structured dictionary for the facilitation of application and learning. Considering that one entity may have several mentions appeared in different positions while not all of them interact with others for relations, we thus adopt the shortest dependency path (SDP) to identify the best mention for building the relation. The dependency parsing tool we use is Spacy\footnote{\url{https://spacy.io/}}. 
\subsection{Statistics and Analysis}
This section conducts some additional statistics and analysis to gain a better understanding of the proposed dataset CEntRE. From the statistics, we have the following observations: 
\begin{itemize}

  \begin{table}[]
    \footnotesize
    \centering
    \begin{tabular}{c|c|c}
    \hline
    Dataset             & \#Relation Types & \#Instance \\ \hline
    SemEval-2010 Task 8 & 10               & 10717      \\ \hline
    FinRE               & 43               & 13700      \\ \hline
    SanWen              & 9                & 15451      \\ \hline
    CEntRE              & 19               & 11018      \\ \hline
    \end{tabular}
    \caption{Size of CEntRE and existing Chinese IE datasets}
    \label{tab:comparison}
    \vspace{-0.4cm}
    \end{table}

\begin{CJK*}{UTF8}{gbsn}
  \begin{table}[]
    \scriptsize
    \centering
    \begin{tabular}{|c|l|}
    \hline
    \multirow{2}{*}{Normal} & \begin{tabular}[c]{@{}l@{}}\textcolor{Red}{字节跳动}则以人民币4.497亿元向\textcolor{Red}{欢喜传媒}购买《囧妈》版权\\。 一切仿佛顺理成章。\end{tabular}
    \\ \cline{2-2} 
                      & \textless{}欢喜传媒，转让，字节跳动\textgreater{}   \\ \hline
    \multirow{2}{*}{SEO}    & \begin{tabular}[c]{@{}l@{}}实际上，\textcolor{Red}{联想}的这些“智能设备”“数据中心集团”，在其竞争\\对手的业务分类里面，或被称之为“消费者业务”（\textcolor{Red}{华为}），\\或者称为“智能手机”（\textcolor{Red}{小米集团}），或者称之为“服务器以\\及微型计算器”（\textcolor{Red}{浪潮信息}），这些才是\textcolor{Red}{联想}业务的实质。\end{tabular}
  
    \\ \cline{2-2} 
                      & \begin{tabular}[c]{@{}l@{}} \textless{}联想，竞争，华为\textgreater{}  \textless{}联想，竞争，小米集团\textgreater{} \\ \textless{}联想，竞争，浪潮信息\textgreater{}\end{tabular}
  
                            \\ \hline
    \multirow{2}{*}{EPO}    & \begin{tabular}[c]{@{}l@{}}2020年美国制裁打压导致\textcolor{Red}{荣耀}手机从\textcolor{Red}{华为}剥离，此后\textcolor{Red}{荣耀}\\将会以竞争对手的形式出现在移动市场。\end{tabular}
  
    \\ \cline{2-2} 
                            & \begin{tabular}[c]{@{}l@{}} \textless{}华为，剥离，荣耀\textgreater{}  \textless{}华为，竞争，荣耀\textgreater{}        \end{tabular}
                            \\ \hline
    \end{tabular}
    \caption{Some annotation examples of the CEntRE}
    \label{data_examples}
    \vspace{-0.4cm}
    \end{table}
  \end{CJK*}

\item[(1)] The current version of the CEntRE dataset contains 11018 labeled triplets and 10894 enterprise entities taking place in 3018 paragraphs. Table~\ref{tab:comparison} shows the comparisons of the CEntRE with its Chinese dataset compeers in scale, where we ignore the unknown relation for the sake of fairness.
\item[(2)] The average number of triplets in each paragraph is 3.65 while the average number of entities is 3.61 reveals the characteristic of overlapping. We divide the dataset into three types according to triplet overlap degree, including Normal, EntityPairOverlap(EPO) and SingleEntityOverlap (SEO). A paragraph belongs to Normal class if none of its triplets have overlapped entities (1241, 11.263\%). A paragraph belongs to EntityPairOverlap class if some of its triplets have overlapped entity pairs (300, 2.723\%). And a paragraph belongs to SingleEntityOverlap class if some of its triplets have an overlapped entity and these triplets do not have overlapped entity pairs (9477, 86.014\%). Table~\ref{data_examples} illustrates all the three types of relation formats: Normal, EPO and SPO. 
\item[(3)] We find that 47.75\%(1441) triplets are associated with more than one sentence, where the longest span is 8 sentences. As the sample(SEO) in Table~\ref{data_examples} shows, some facts could only be extracted from multi-sentences, which demonstrates that the dataset is a good benchmark for RE reasoning. We can also conclude that a perfect model should have rich abilities of reading, synthesizing, and reasoning information for CEntRE. 
\item[(4)] Unfortunately, there are unbalanced amounts in different relations with the maximum number of triplets 2095 in \begin{CJK*}{UTF8}{gbsn}“附属”\end{CJK*} but the minimum number 26 in \begin{CJK*}{UTF8}{gbsn}“代管”\end{CJK*}, which may be due to the nature of commercial activities. The unbalanced proportion aggregates the difficulty of relations classification and we will improve it in the future.
\end{itemize}

\section{Experiments and Analysis}
\label{sec:experiments}
In this section, we introduce the baseline models, experimental settings and evaluation metrics.

\subsection{Baseline Models}
There are two kinds of approaches for the NRE task \cite{14}: the pipelined framework, which first uses named entity recognition (NER) models to detect entity spans, and then implements relation extraction between pairs of detected spans; and the joint learning method, which combines the NER model and 
\begin{table}[htbp]
  \centering
  \scriptsize
  \setlength{\tabcolsep}{2mm}
  \begin{tabular}{|cc|ccc|ccc|}
  \hline
  \multicolumn{2}{|c|}{Models}                                                                                   & \multicolumn{3}{c|}{NER}                                        & \multicolumn{3}{c|}{RE}                                         \\ \hline
  \multicolumn{2}{|c|}{\textbackslash{}}                                                                         & \multicolumn{1}{c|}{P}     & \multicolumn{1}{c|}{R}     & F1    & \multicolumn{1}{c|}{P}     & \multicolumn{1}{c|}{R}     & F1    \\ \hline
  \multicolumn{1}{|c|}{\multirow{2}{*}{\begin{tabular}[c]{@{}c@{}}Pipe\\ Works\end{tabular}}}  & LSTM+MLP       & \multicolumn{1}{c|}{89.52} & \multicolumn{1}{c|}{81.01} & 85.05 & \multicolumn{1}{c|}{44.32} & \multicolumn{1}{c|}{42.89} & 43.59 \\ \cline{2-8} 
  \multicolumn{1}{|c|}{}                                                                        & BERT+MLP       & \multicolumn{1}{c|}{93.33} & \multicolumn{1}{c|}{85.61} & 89.30 & \multicolumn{1}{c|}{46.05} & \multicolumn{1}{c|}{45.61} & 45.83 \\ \hline
  \multicolumn{1}{|c|}{}                                                                        & Spert          & \multicolumn{1}{c|}{93.60} & \multicolumn{1}{c|}{87.12} & 90.24 & \multicolumn{1}{c|}{46.13} & \multicolumn{1}{c|}{51.28} & 48.57 \\ \cline{2-8} 
  \multicolumn{1}{|c|}{\multirow{2}{*}{\begin{tabular}[c]{@{}c@{}}Joint\\ Works\end{tabular}}} & CasRel         & \multicolumn{1}{c|}{94.69} & \multicolumn{1}{c|}{86.31} & 90.31 & \multicolumn{1}{c|}{48.28} & \multicolumn{1}{c|}{45.89} & 47.05 \\ \cline{2-8} 
  \multicolumn{1}{|c|}{}                                                                        & TP-Linker      & \multicolumn{1}{c|}{89.97} & \multicolumn{1}{c|}{88.73} & 89.35 & \multicolumn{1}{c|}{47.27} & \multicolumn{1}{c|}{46.92} & 47.09 \\ \cline{2-8} 
  \multicolumn{1}{|c|}{}                                                                        & Table-Sequence & \multicolumn{1}{c|}{91.37} & \multicolumn{1}{c|}{89.02} & 91.18 & \multicolumn{1}{c|}{49.24} & \multicolumn{1}{c|}{47.65} & 48.43 \\ \hline
  \end{tabular}
  \caption{Performance comparison on CEntRE}
  \label{baselines}
  \vspace{-0.4cm}
  \end{table}
  the RE model through different strategies, such as constraints or parameters sharing.

By these works, we selected six representative models covering the pipeline work (LSTM+MLP, BERT+MLP) and the joint learning (Table-Sequence\cite{TwoThanOne}, CasRel\cite{casrel}, TP-Linker\cite{TPLinker}, Spert\cite{41}) to conduct experiments on our dataset.

\subsection{Experimental Setting and Metrics}
\textbf{Setting:} We use the ratio of 8:1:1 to divide the entire dataset into training set, test set and validation set. Each model was trained for 10 times with 40 epochs each time, and we report the average result on the test set. As for the other hyperparameters, we keep all them unchanged.

\textbf{Metrics:} We evaluate NER and RE with Precision (P), Recall (R), and micro F1 scores, while a predicted entity is correct if its boundaries are correct and a predicted relation is true if the triplet (subject, relation, object) is infallible.  

\subsection{Results and Discussions}

All experimental results are summarized in Table \ref{baselines}, and there are several important observations from the table. (1) Comparing LSTM+MLP and BERT+MLP, we find that the performances of BERT+MLP are better than LSTM+MLP on both tasks. This indicates that the pre-trained model is quite effective due to its large external knowledge support. (2) It is clear that Joint-based methods perform better compared with all of the Pipe-based works since such pipeline models cannot capture the explicit interaction between the two sub-tasks and limits the performance of each component. (3) The Spert can get the best RE recall among all the models, but it costs much time to enumerate all span-pairs for predicting the relations, which is unreachable for actual deployment and application. (4) The RE effects of all models are far from the human annotation, highlighting the necessitation to identify effective mentions and capture the inter-sentence context. This reflects the challenges of CEnterRE and demonstrates that these models are not sufficient to evaluate the complexity of this dataset, presenting the opportunities for the future work. 

\section{Conclusion}
In this paper, we present a new Chinese dataset CEntRE with large data volume and high coverage. Our dataset focuses on enterprise relation extraction, and we hope it can prompt the research in related areas. We then conduct extensive evaluation on CEntRE with several excellent models, and experimental results indicate the challenges for the future research.

\bibliographystyle{IEEEbib}
\bibliography{strings,refs}

\begin{thebibliography}{10}

\bibitem{12}
Daojian Zeng, Haoran Zhang, and Qianying Liu,
\newblock ``Copymtl: Copy mechanism for joint extraction of entities and
  relations with multi-task learning,''
\newblock {\em Proceedings of the AAAI Conference on Artificial Intelligence},
  vol. 34, no. 05, pp. 9507--9514, Apr. 2020.

\bibitem{YangCL}
Chuanlong Yang and Jinlong. Wang,
\newblock ``Research on automatic extraction of enterprise supply relationship
  based on nlp,''
\newblock {\em Computer Science and Application}, vol. 008, no. 012, pp.
  1823--1832, November 2018.

\bibitem{MengL}
Lei Meng, Zhi-Hao Wei, Tian-Yi Hu, Qian Cheng, Yu~Zhu, and Xiao-Tao Wei,
\newblock ``An improved method for chinese company name and abbreviation
  recognition,''
\newblock in {\em International Conference on Knowledge Management in
  Organizations}, Lorna Uden, Wei Lu, and I-Hsien Ting, Eds., Cham, 2017, pp.
  435--447, Springer International Publishing.

\bibitem{DaiJB}
Jiangbo Dai, Jianhua Mao, Xuefeng Liu, and et~al,
\newblock ``Enterprise ecological relationship extraction based on feature
  vector and svo extension,''
\newblock {\em Computer Technology and Development}, vol. 28, no. 10, pp.
  146--151, October 2018.

\bibitem{SunC}
Chen Sun, Yingnan Fu, Wenliang Cheng, Wei-ning Qian, and ect,
\newblock ``Chinese named entity relation extraction for enterprise knowledge
  graph construction,''
\newblock {\em Journal of East China Normal University (Natural Science)}, vol.
  199, no. 03, pp. 60--71, May 2018.

\bibitem{13}
Weipeng Huang, Xingyi Cheng, Taifeng Wang, and Wei Chu,
\newblock ``Bert-based multi-head selection for joint entity-relation
  extraction,''
\newblock in {\em Natural Language Processing and Chinese Computing}, Jie Tang,
  Min-Yen Kan, Dongyan Zhao, Sujian Li, and Hongying Zan, Eds., Cham, 2019, pp.
  713--723, Springer International Publishing.

\bibitem{14}
Kui Xue, Yangming Zhou, Zhiyuan Ma, Tong Ruan, Huanhuan Zhang, and Ping He,
\newblock ``Fine-tuning bert for joint entity and relation extraction in
  chinese medical text,''
\newblock in {\em 2019 IEEE International Conference on Bioinformatics and
  Biomedicine (BIBM)}, 2019, pp. 892--897.

\bibitem{39}
Giannis Bekoulis, Johannes Deleu, Thomas Demeester, and Chris Develder,
\newblock ``Joint entity recognition and relation extraction as a multi-head
  selection problem,''
\newblock {\em Expert Systems with Applications}, vol. 114, pp. 34--45, 2018.

\bibitem{41}
M.~Eberts and A.~Ulges,
\newblock ``Span-based joint entity and relation extraction with transformer
  pre-training,''
\newblock {\em arXiv}, 2019.

\bibitem{ChenDQ}
Zexuan Zhong and Danqi Chen,
\newblock ``A frustratingly easy approach for entity and relation extraction,''
\newblock in {\em Proceedings of the 2021 Conference of the North American
  Chapter of the Association for Computational Linguistics: Human Language
  Technologies}, Online, June 2021, pp. 50--61, Association for Computational
  Linguistics.

\bibitem{Dai}
Dai Dai, Xinyan Xiao, Yajuan Lyu, Shan Dou, Qiaoqiao She, and Haifeng Wang,
\newblock ``Joint extraction of entities and overlapping relations using
  position-attentive sequence labeling,''
\newblock {\em Proceedings of the AAAI Conference on Artificial Intelligence},
  vol. 33, no. 01, pp. 6300--6308, Jul. 2019.

\bibitem{TwoThanOne}
Jue Wang and Wei Lu,
\newblock ``Two are better than one: Joint entity and relation extraction with
  table-sequence encoders,''
\newblock in {\em Proceedings of the 2020 Conference on Empirical Methods in
  Natural Language Processing (EMNLP)}, Online, Nov. 2020, pp. 1706--1721,
  Association for Computational Linguistics.

\bibitem{8}
Tong Ruan, Lijuan Xue, Haofen Wang, Fanghuai Hu, Liang Zhao, and Jun Ding,
\newblock ``Building and exploring an enterprise knowledge graph for investment
  analysis,''
\newblock in {\em The Semantic Web -- ISWC 2016}, Paul Groth, Elena Simperl,
  Alasdair Gray, Marta Sabou, Markus Kr{\"o}tzsch, Freddy Lecue, Fabian
  Fl{\"o}ck, and Yolanda Gil, Eds., Cham, 2016, pp. 418--436, Springer
  International Publishing.

\bibitem{11}
Ziran Li, Ning Ding, Zhiyuan Liu, Haitao Zheng, and Ying Shen,
\newblock ``{C}hinese relation extraction with multi-grained information and
  external linguistic knowledge,''
\newblock in {\em Proceedings of the 57th Annual Meeting of the Association for
  Computational Linguistics}, Florence, Italy, July 2019, pp. 4377--4386,
  Association for Computational Linguistics.

\bibitem{casrel}
Zhepei Wei, Jianlin Su, Yue Wang, Yuan Tian, and Yi~Chang,
\newblock ``A novel cascade binary tagging framework for relational triple
  extraction,''
\newblock in {\em Proceedings of the 58th Annual Meeting of the Association for
  Computational Linguistics}, Online, July 2020, pp. 1476--1488, Association
  for Computational Linguistics.

\bibitem{TPLinker}
Yucheng Wang, Bowen Yu, Yueyang Zhang, Tingwen Liu, Hongsong Zhu, and Limin
  Sun,
\newblock ``Tplinker: Single-stage joint extraction of entities and relations
  through token pair linking,''
\newblock 01 2020, pp. 1572--1582.

\end{thebibliography}

\end{document}